\documentclass[letterpaper, 10 pt, conference]{ieeeconf}  
\let\labelindent\relax

\usepackage{graphicx}
\usepackage{amsfonts}
\usepackage{amsmath,amssymb,bm}
\usepackage{upgreek}
\usepackage{multirow}
\usepackage{booktabs}
\usepackage{enumitem}
\usepackage{caption}
\usepackage{cite}
\usepackage{amsmath}  
\usepackage{makecell}     
\usepackage{multirow}     

\IEEEoverridecommandlockouts               
\title{\LARGE \bf
IONext: Unlocking the Next Era of Inertial Odometry
}

\author{
Shanshan Zhang$^{1,2}$, Qi Zhang$^{1}$, Siyue Wang$^{1}$, Tianshui Wen$^{1}$, Liqin Wu$^{1}$, Ziheng Zhou$^{1}$, \\ Xuemin Hong$^{1}$, Ao Peng$^{1}$, Lingxiang Zheng$^{1,*}$, Yu Yang$^{2,*}$
\thanks{This work is supported by Science and Technology Major Program of Fujian Province (No. 2022HZ026007) and partly supported by the Science and Technology Planning Project of Fujian province (2022I0001).}
\thanks{$^{1}$Shanshan Zhang, Qi Zhang, Siyue Wang, Tianshui Wen, Ao Peng, Xuemin Hong and Lingxiang Zheng are with the Department of Information and Communication Engineering, National and Local Joint Engineering Research Center of Navigation and Location-Based Services, Xiamen University, Xiamen 361005, China (e-mail: lxzheng@xmu.edu.cn).}
\thanks{$^{2}$Yu Yang and Shanshan Zhang are with the Department of Electronic Science, State Key Laboratory of Physical Chemistry of Solid Surfaces, Xiamen University, Xiamen 361005, China (e-mail: yuyang15@xmu.edu.cn).}
}

\begin{document}

\maketitle
\thispagestyle{empty}
\pagestyle{empty}

\begin{abstract}
Attention mechanisms have recently achieved notable success in inertial odometry (IO). However, their limited sensitivity to local, fine-grained motion variations and lack of inherent inductive biases often constrain localization accuracy and generalization. Conversely, prior studies have shown that augmenting CNNs with \textit{large-$k$ convolutions} and Transformer-inspired architectural designs can effectively expand the receptive field and achieve performance comparable to attention-based methods. Motivated by these insights, we propose the Adaptive Dynamic Mixer (ADM), which first employs multi-scale convolutional kernels to extract both contextual motion information and fine-grained local motion features, and then generates dynamic weights from the input to adaptively aggregate multi-scale features while preserving convolutional inductive biases. We further introduce an Adaptive Gating Unit (AGU) to improve cross-channel modeling and adaptively regulate the features of nearest-neighbor motion variations. Building on ADM and AGU, and using the Transformer as a structural reference, we develop a CNN-based IO backbone, IONext. Extensive experiments on six public datasets demonstrate that IONext consistently outperforms existing IO methods, achieving state-of-the-art performance. To adhere to the double-blind review requirements, the code will be released to the public following the conclusion of the peer review.
\end{abstract}

\section{INTRODUCTION}
Inertial odometry (IO) aims to accurately estimate motion using accelerometer and gyroscope measurements from an inertial measurement unit (IMU)~\cite{AirIO}. This approach requires no additional external hardware beyond the IMU and operates independently of external infrastructure or environmental conditions, making it well-suited for infrastructure-free localization in civilian applications~\cite{Tartan-IMU,RIO}.

Before the widespread adoption of machine learning for IO, researchers primarily relied on Newtonian mechanics to estimate motion states from IMU measurements\cite{SurveyofIndoorInertial}. However, IMU measurements inevitably contain noise, whose accumulation leads to severe drift in traditional double integration methods\cite{SINS}. Prior work has incorporated physical priors to mitigate drift\cite{PDRusingfrequencydomain,AdaptiveThreshold-BasedZUPT,IMU-and-magnetometer-modeling-for-smartphone-based-PDR}; however, such priors often restrict admissible motion states and reduce adaptability in challenging environments\cite{RoNIN}.

The advent of data-driven methods shifted IO toward learning-based approaches that infer motion patterns from large volumes of IMU data, improving robustness to adverse conditions and measurement noise\cite{RoNIN}. 
For example, CNN-based IO methods exploit the inductive bias of convolution (locality, translation invariance) to enhance their ability to capture fine-grained motion variations, laying a foundation for handling structured motion patterns\cite{IMUNet,RoNIN}. 
However, CNNs have limited receptive fields and thus struggle to model contextual motion information\cite{InceptionTransformer}. Inspired by the success of attention in Natural Language Processing (NLP)\cite{Attentionisallyouneed} and Computer Vision (CV)\cite{TKSA}, Transformer-based methods have been applied to IO to capture contextual motion information and improve localization accuracy\cite{DeepILS,CTIN,iMOT}. Nevertheless, these methods may reduce sensitivity to fine-grained motion variations and lack the inductive bias of convolution, which can degrade generalization.

Outside IO, studies have demonstrated two relevant findings: (1) integrating Transformer-style design principles into CNNs can yield performance comparable to that of Transformers\cite{ConvNext}, and (2) \textit{large-$k$ convolutions} can approximate a Transformer's receptive field while preserving the inductive bias of convolution\cite{RepLKNet,AlexNet,Inceptionv1}. These insights are particularly relevant for IO, where both fine-grained motion variations and long-range dependencies must be modeled simultaneously.

\begin{figure}[t]
\centering
\captionsetup{aboveskip=2pt,font=small}
\includegraphics[width=0.48\textwidth]{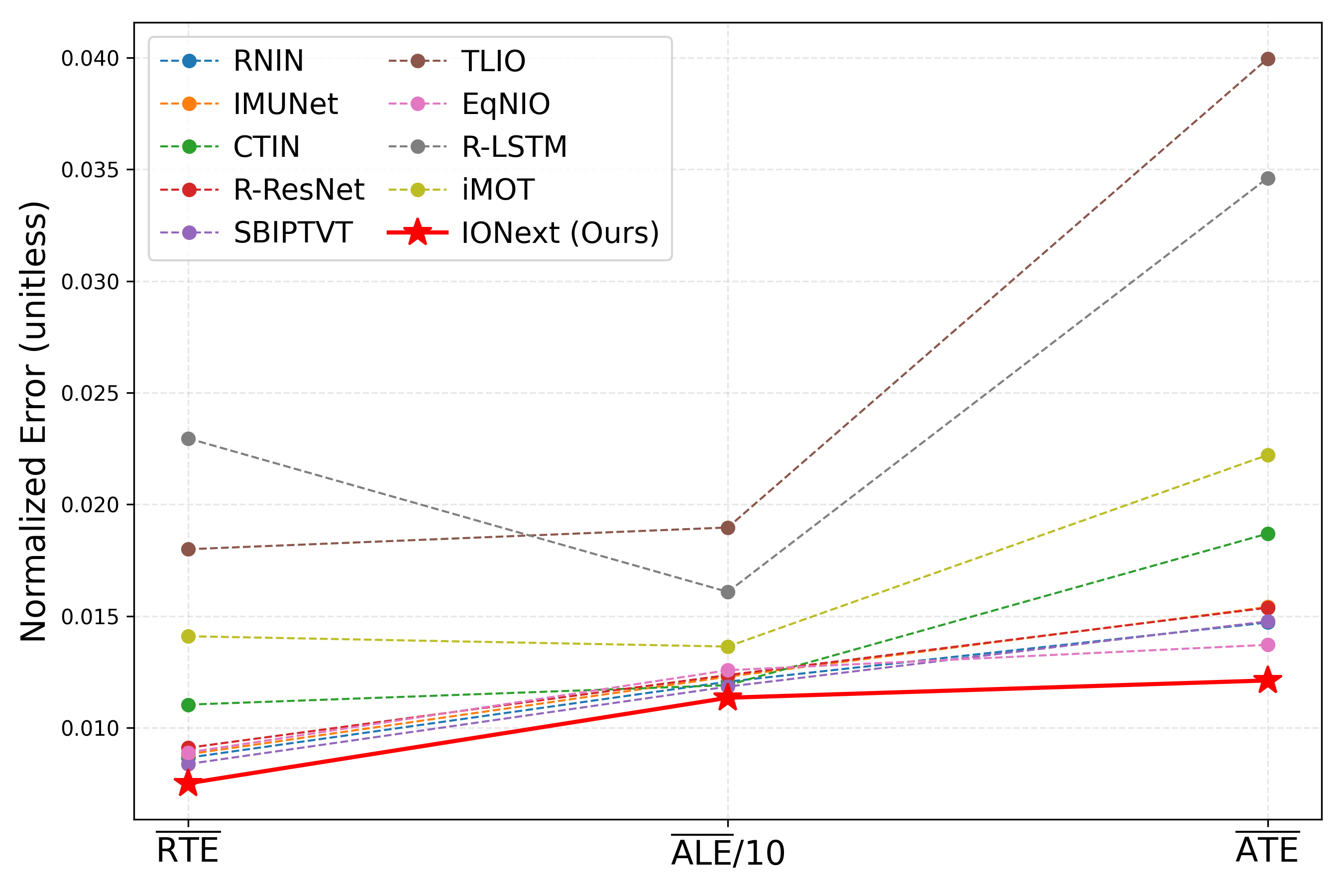}
\caption{Comparison of $\overline{ATE}$, $\overline{RTE}$, and $\overline{ALE}$ on the RNIN dataset. IONext achieves lower errors than baselines.}
\label{model_comparison}
\vspace{-20pt}
\end{figure}

Nevertheless, naively enlarging convolutional kernels reduces sensitivity to fine-grained motion\cite{Inceptionnext}. Moreover, although convolutional inductive bias supports generalization, fixed convolutional parameters cannot adapt to dynamic IMU inputs. This difference in input-processing underlies the gap between convolutional and attention-based IO\cite{TransXNet}. Therefore, a key question is how to design an IO model that simultaneously extracts local fine-grained motion features and contextual motion information, preserves the beneficial inductive bias of convolution, and adapts to dynamic IMU inputs.

To address these challenges, we propose IONext. Extensive experiments verify its effectiveness in IO tasks, as shown in Fig.~\ref{model_comparison}.
Specifically, our contributions are as follows:
\begin{itemize}[noitemsep, nolistsep, leftmargin=*]
  \item We propose the Adaptive Dynamic Mixer (ADM), a module that adaptively processes inputs to extract local fine-grained motion variations and to model contextual motion information while preserving the inductive bias of convolution.
  \item We present the Adaptive Gating Unit (AGU), which enhances channel modeling capability and adaptively regulates nearest-neighbor motion-variation features.
  \item Based on ADM and AGU and inspired by Transformer design principles, we propose the convolutional architecture IONext.
  \item We introduce the Absolute Length Error (ALE), a new evaluation metric that addresses the lack of trajectory-length-aware evaluation. We also propose a length-based normalization strategy to remove bias introduced by differing trajectory lengths.
\end{itemize}

\section{Related Work}
\subsection{Data-Driven IO Methods}
Data-driven IO methods have substantially broadened the application scope of IO, reducing sensitivity to device placement and to specific motion patterns.

\textbf{Pre-Transformer:} RIDI~\cite{RIDI} and PDRNet~\cite{PDRNet} first classify device placement and then train specialized networks for velocity inference. By contrast, IONet~\cite{Ionet} and RoNIN~\cite{RoNIN} adopt unified deep architectures for velocity estimation and demonstrate strong generalization. RoNIN~\cite{RoNIN} evaluates several backbones---including ResNet, TCN, and LSTM---and finds that ResNet, as a convolutional backbone, is particularly effective at extracting fine-grained motion features.

To improve CNN performance, researchers have proposed several strategies: TLIO~\cite{TLIO} and LIDR~\cite{LIDR} apply filter--based post-processing to refine ResNet outputs; WDSNet~\cite{WDSNet} uses wavelet-based signal selection to enhance input quality; IMUNet~\cite{IMUNet} adopts depthwise-separable convolutions for lightweight mobile deployment; and RIO~\cite{RIO} and EqNIO~\cite{EqNIO} exploit motion equivariance and modular components to improve adaptability and accuracy.

Although these approaches advance CNN-based IO from multiple perspectives, they still struggle to capture contextual motion information. Combining CNNs with RNNs partially mitigates this limitation, but fixed convolutional parameters remain unable to adapt dynamically to changing IMU measurements.

\textbf{Transformer:} Originally devised for natural language processing (NLP), attention mechanisms have been successfully transferred to computer vision (CV) and multimodal tasks. Motivated by these advances, researchers have applied attention to IO: DeepILS~\cite{DeepILS} and NLOC~\cite{NILoc} introduce attention modules; SBIPTVT~\cite{SBIPTVT} employs a Transformer encoder for real-time pedestrian velocity estimation; CTIN~\cite{CTIN} and iMOT~\cite{iMOT} develop full encoder--decoder architectures, with CTIN incorporating temporal embeddings and iMOT using a particle-initialization mechanism.

Transformer-based methods excel at modeling global dependencies and can dynamically compute attention matrices conditioned on IMU measurements. However, they are less effective at capturing fine-grained motion variations and lack the convolutional inductive bias, which can limit generalization\cite{TransXNet}.

\subsection{Beneficial Explorations}
CNNs and Transformers offer complementary strengths in inductive bias and dynamic modeling, which motivates hybrid approaches. 
A straightforward direction is to inject convolutional inductive bias into attention mechanisms. However, modifying attention is challenging, and the quadratic complexity of attention matrices places a substantial computational burden on mobile devices. For example, Swin Transformer~\cite{Swin-ransformer-V2} employs shifted-window self-attention to retain some inductive bias, but its receptive field remains limited.
Recent CNN improvements follow two main directions:
\begin{itemize}[noitemsep, nolistsep, leftmargin=*]
  \item \textbf{Expanding the receptive field via large kernels:} Early models (e.g., AlexNet~\cite{AlexNet}, InceptionV1~\cite{Inceptionv1}) adopted large kernels (e.g., $11\times11$, $7\times7$) to enlarge the receptive field. To reduce computational cost, later architectures (e.g., InceptionV3~\cite{Inceptionv3}, SLaK~\cite{SLaK}) decompose large kernels into parallel branches to capture contextual and local features simultaneously.
  \item \textbf{Designing Transformer-like CNNs:} By adopting Transformer design principles and training techniques, CNNs have achieved substantial gains. For instance, replacing attention in Swin Transformer with dynamic depthwise convolutions preserves accuracy~\cite{Local-Attention-and-Dynamic-Depth-wise-Convolution}; ConvNeXt~\cite{ConvNext} progressively integrates Transformer design choices and outperforms the Swin Transformer on several vision benchmarks.
\end{itemize}

Despite these advances, improving the adaptability of CNNs to dynamically varying inputs has received relatively little attention. To address this gap, we propose IONext, which adaptively processes inputs to extract local fine-grained motion variations while modeling contextual motion and preserving the convolutional inductive bias.

\section{Method and Architecture}
In this section, we present the overall framework of the proposed IONext and describe its core component, the Adaptive Dynamic Encoder (ADE), which comprises the ADM and the AGU.

\subsection{Architecture Design}
\begin{figure*}[!t]
\centering
\captionsetup{aboveskip=2pt,font=small}
\includegraphics[width=0.95\textwidth]{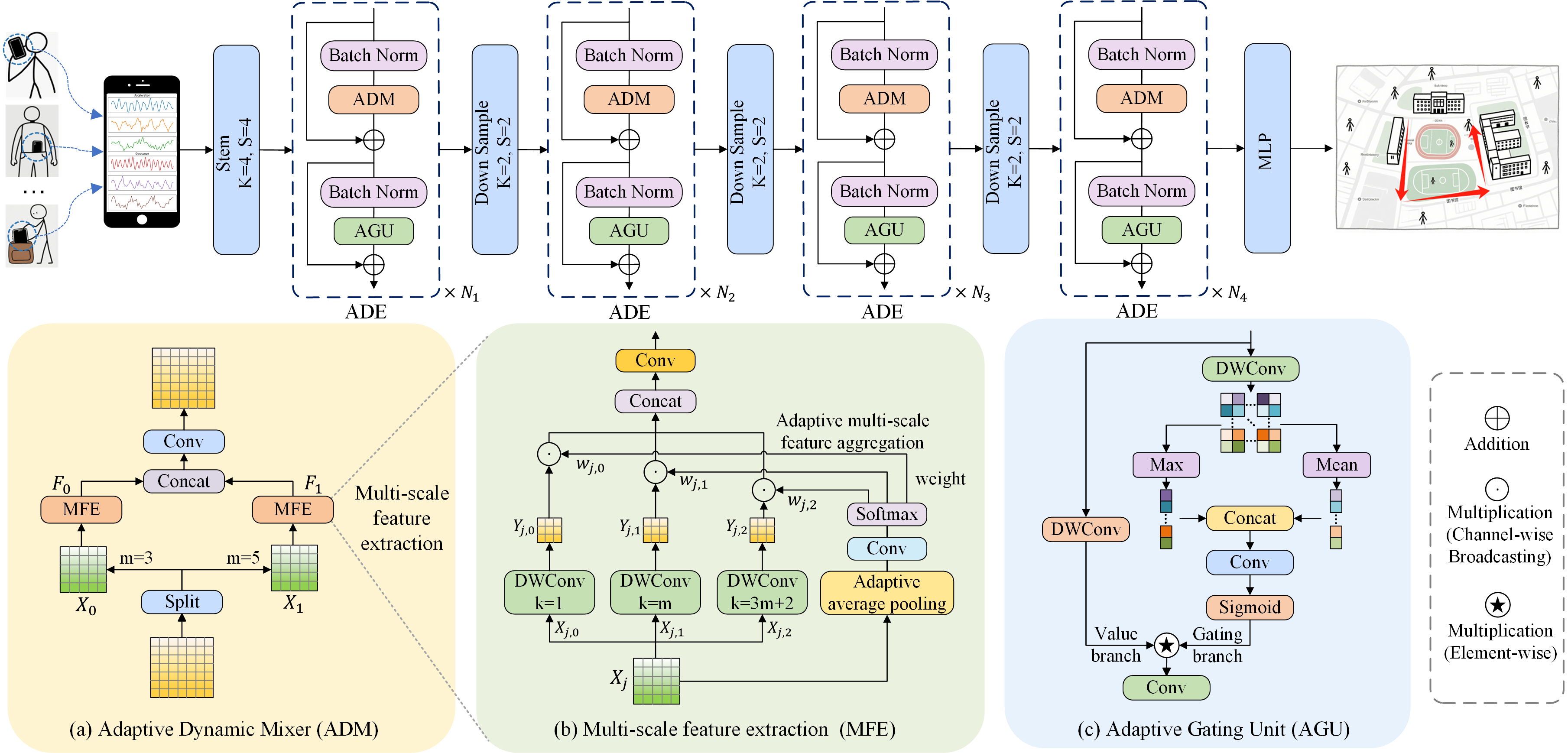}
\caption{The overall architecture of the proposed IONext consists of the ADE, which comprises the ADM and the AGU.}
\label{framework}
\vspace{-20pt}
\end{figure*}

The overall structure of IONext is shown in Fig.~\ref{framework}. It is a CNN-based architecture tailored for IO, which incorporates several structural design principles from the Swin Transformer\cite{Swin-ransformer-V2}---including network depth, channel widths, stem and downsampling layers---to enhance model accuracy.

Given IMU measurements \(\mathbf{X}\in\mathbb{R}^{C\times T}\) over a one-second window (\(1~\mathrm{s}\)), where \(C=6\) denotes the six channels (triaxial accelerometer and triaxial gyroscope) and \(T\) denotes the window length (equal to the sampling frequency in samples per second), the matrix \(\mathbf{X}\) is first downsampled via a 1D non-overlapping convolution to produce encoder-ready token sequences. Because non-overlapping convolution preserves the relative temporal structure of motion signals, explicit positional encoding is not required.

The backbone of IONext, maintaining structural consistency with the Swin-Transformer\cite{Swin-ransformer-V2}, comprises four stages with \(N_i=[2,2,6,2]\) stacked ADE blocks for multi-scale feature extraction. To enhance representational capacity for noisy IMU data, each stage employs different channel widths to facilitate effective multi-scale modeling. Feature downsampling is performed using non-overlapping convolutions with kernel size \(k=2\) and stride \(s=2\).

The network output is the average velocity over the unit time window, and trajectories are reconstructed by integrating the predicted velocity sequence. During training, the model is optimized by minimizing the mean squared error (MSE) loss.

\subsection{Adaptive Dynamic Encoder}
IMU measurements inherently contain multi-scale motion information, ranging from fine-grained motion variations (e.g., sudden turns) to contextual motion information (e.g., steady walking). To simultaneously model this rich motion information, we propose the ADM and AGU modules, which together constitute the ADE. This encoder replaces the traditional Transformer encoder, preserving the inductive biases of convolutional networks while enabling multi-scale feature extraction.

Concretely, consider the input to the $n$-th encoder block as $\mathbf{X_n} \in \mathbb{R}^{C \times T}$, where $C$ denotes the channel dimension and $T$ the temporal length. The computation within the ADE is defined as:
\begin{equation}
\mathbf{X_n'} = \mathbf{X_n} + ADM(BN(\mathbf{X_n}))
\end{equation}
\begin{equation}
\mathbf{X_{n+1}} = \mathbf{X_n'} + AGU(BN(\mathbf{X_n'}))
\end{equation}
where $BN(\cdot)$ denotes batch normalization along the channel dimension, and $\mathbf{X_{n+1}}$ represent the outputs of the $n$-th ADE module. The detailed architectures of ADM and AGU are described in the following sections.

\textbf{Adaptive Dynamic Mixer (ADM).} We first briefly review the standard self-attention mechanism. Given input tokens $\mathbf{X_n} \in \mathbb{R}^{C \times T}$, queries $Q$, keys $K$, and values $V$ are obtained via linear projections of $\mathbf{X_n}$. The standard self-attention process can then be expressed as:
\begin{equation}
    \text{Self-Attention}(Q, K, V) = \operatorname{softmax}\left( \frac{QK^\top}{\sqrt{T}} \right) V
\end{equation}
This mechanism enables modeling of contextual motion information and dynamically computes the attention matrix based on the input data. However, it is computationally expensive and compromises the inductive biases of CNNs as well as their ability to capture fine-grained motion variations.

\begin{table*}[!t]
\centering
\captionsetup{aboveskip=2pt,font=small}
\scriptsize
\renewcommand{\arraystretch}{0.9}
\caption{Comparison of normalized error rankings on six datasets. Bold and underline indicate the best and second-best results. To enhance data readability, all indicators have been scaled up by a factor of 100.}
\label{results}
\begin{tabular}{llccccccccccc}
        \toprule
        \multicolumn{3}{c}{\textbf{Model Classification}} 
        & \multicolumn{5}{c}{\textbf{CNN-based}} 
        & \multicolumn{1}{c}{\textbf{LSTM-based}} 
        & \multicolumn{2}{c}{\textbf{Hybrid}} 
        & \multicolumn{2}{c}{\textbf{Transformer-based}} \\
        \cmidrule(lr){4-8} \cmidrule(lr){9-9} \cmidrule(lr){10-11} \cmidrule(lr){12-13}

        \multicolumn{3}{c}{\textbf{Models}} 
        & IONext & RoNIN ResNet & TLIO & IMUNet 
        & EqNIO & RoNIN LSTM & RNIN & SBIPTVTL & CTIN & iMOT \\
        \midrule

        \multicolumn{3}{c}{\textbf{Publication}} 
        & \makecell{-\\-} 
        & \makecell{ICRA\\2020} 
        & \makecell{RA-L\\2020} 
        & \makecell{TIM\\2024} 
        & \makecell{ICLR\\2025} 
        & \makecell{ICRA\\2020} 
        & \makecell{ISMAR\\2021} 
        & \makecell{CSCWD\\2024} 
        & \makecell{AAAI\\2022} 
        & \makecell{AAAI\\2025} \\
        \midrule

        \multirow{3}{*}{\textbf{RIDI}} 
        &$\overline{ATE}$ 
        & \multirow{3}{*}{$(\times10^{-2})$} & \textbf{1.41} & 1.66 & 1.70 & 1.52 & 1.53 
        & 3.13 & 1.78 & \underline{1.46} & 1.66 & 1.93 \\
        &$\overline{RTE}$ 
        &  & \underline{1.71} & 1.82 & 1.95 & 1.86 & 1.72 
        & 2.72 & 2.09 & \textbf{1.65} & 1.92 & 2.30 \\
        &$\overline{ALE}$
        &  & \textbf{2.19} & 3.84 & 4.87 & \underline{2.27} & 4.06 
        & 3.11 & 3.93 & 2.37 & 3.41 & 3.15 \\
        \midrule

        \multirow{3}{*}{\textbf{RoNIN}} 
        &$\overline{ATE}$  
        & \multirow{3}{*}{$(\times10^{-2})$} & \underline{1.03} & 1.09 & 1.42 & 1.29 & \textbf{0.99} 
        & 1.98 & 1.20 & 1.10 & 1.16 & 1.23 \\
        &$\overline{RTE}$ 
        &  & \underline{0.92} & 0.97 & 1.06 & 0.99 & \textbf{0.89} 
        & 1.27 & 0.96 & 0.95 & 1.00 & 1.06 \\
        &$\overline{ALE}$
        &  & \textbf{4.44} & 5.63 & 5.76 & 5.84 & \underline{5.24} 
        & 13.05 & 5.40 & 6.15 & 6.12 & 8.32 \\
        \midrule

        \multirow{3}{*}{\textbf{TLIO}} 
        &$\overline{ATE}$  
        & \multirow{3}{*}{$(\times10^{-2})$} & \textbf{0.86} & 1.01 & 1.07 & 2.14 & \underline{0.92} 
        & 2.42 & 1.05 & 1.08 & 1.71 & 2.04 \\
        &$\overline{RTE}$
        &  & \textbf{0.51} & 0.61 & 0.66 & 0.85 & \underline{0.56} 
        & 1.45 & 0.61 & 0.68 & 0.78 & 0.97 \\
        &$\overline{ALE}$ 
        &  & \textbf{2.59} & 4.98 & 4.33 & \underline{3.14} & 3.65 
        & 13.89 & 3.33 & 3.83 & 3.33 & 5.68 \\
        \midrule
        
        \multirow{3}{*}{\textbf{RNIN}} 
        &$\overline{ATE}$ 
        & \multirow{3}{*}{$(\times10^{-2})$} & \textbf{1.21} & 1.54 & 4.00 & 1.54 & \underline{1.37} 
        & 3.46 & 1.47 & 1.48 & 1.87 & 2.22 \\
        &$\overline{RTE}$
        &  & \textbf{0.75} & 0.91 & 1.80 & 0.88 & 0.89 
        & 2.30 & 0.87 & \underline{0.84} & 1.10 & 1.41 \\
        &$\overline{ALE}$
        &  & \textbf{11.35} & 12.36 & 18.97 & 12.29 & 12.58 
        & 16.08 & 12.07 & \underline{11.84} & 11.91 & 13.64 \\
        \midrule

        \multirow{3}{*}{\textbf{IMUNet}} 
        &$\overline{ATE}$  
        & \multirow{3}{*}{$(\times10^{-2})$} & \textbf{2.22} & 2.76 & 3.67 & 2.84 & 3.44 
        & 3.57 & 2.70 & 3.25 & 2.72 & \underline{2.35} \\
        &$\overline{RTE}$
        &  & \textbf{1.60} & 1.98 & 2.22 & 1.97 & 2.92 
        & 2.53 & 1.96 & 2.32 & 1.99 & \underline{1.80} \\
        &$\overline{ALE}$
        &  & \textbf{5.79} & 9.62 & 9.19 & 6.39 & 7.02 
        & 14.96 & 8.06 & 9.36 &  \underline{6.22} &6.99 \\
        \midrule

        \multirow{3}{*}{\textbf{OxIOD}} 
        &$\overline{ATE}$ 
        & \multirow{3}{*}{$(\times10^{-2})$} & \textbf{0.49} & 0.54 & 0.71 & 0.65 & 0.55 
        & 5.54 & 0.61 & \underline{0.50} & 1.06 & 1.26 \\
        &$\overline{RTE}$
        &  & \textbf{0.38} & \textbf{0.38} & 0.40 & \textbf{0.38} & \underline{0.39}
        & 1.40 & 0.42 & \underline{0.39} & 0.48 & 0.60 \\
        &$\overline{ALE}$
        &  & \textbf{5.16} & 5.77 & 7.17 & 6.44 & 6.58 
        & 32.58 & 6.27 & \underline{5.64} & 7.21 & 11.38 \\
        \bottomrule
    \end{tabular}%
    \vspace{-10pt}
\end{table*}

\begin{figure*}[!t]
\centering
\captionsetup{aboveskip=2pt,font=small}
\includegraphics[width=0.98\textwidth]{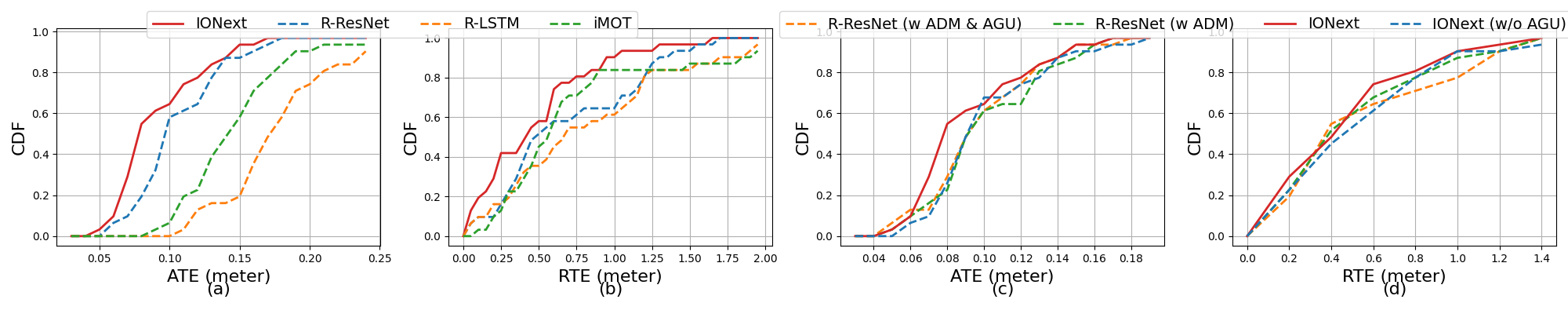}
\caption{Performance evaluation on the RNIN dataset. (a)–(b): ATE/RTE CDF curves of IONext vs. baselines. (c)–(d): Effects of adding modules to IONext and R-ResNet. Curves closer to the top-left indicate better performance.}
\label{CDF_and_PDE}
\vspace{-20pt}
\end{figure*}

To overcome these limitations, we propose the ADM structure, which integrates multi-scale convolution with an adaptive feature aggregation mechanism for more efficient multi-scale modeling. The input $\mathbf{X_n}$ is first evenly split along the channel dimension into two sub-tensors: $X_j \in \mathbb{R}^{\frac{C}{2} \times T}$, $j \in \{0, 1\}$. Each sub-tensor undergoes multi-scale feature extraction (MFE) through a set of parallel convolutions, as shown in Fig.~2(b). The MFE module contains three parallel 1D depthwise convolutions with kernel sizes of $1$, $k$, and $3k+2$, where $k \in \{3,5\}$. These branches extract features at different scales to capture both contextual motion information and fine-grained motion variations. The processing of an individual branch is expressed as:
\begin{equation}
    Y_{j,i} = \operatorname{DWConv}_i(\mathbf{X}_j), \quad i \in \{0,1,2\}
\end{equation}
where $\operatorname{DWConv}_i$ denotes the 1D depthwise convolution with kernel sizes corresponding to $1 (i=0)$, $k (i=1)$, and $3k+2 (i=2)$. This design preserves CNN inductive biases while extending the receptive field for contextual information.

To adaptively fuse multi-scale feature maps, we introduce an input-dependent weighting mechanism\cite{TransXNet}. For each $X_j$, we first apply adaptive global average pooling along the temporal dimension $T$ to obtain channel-wise statistics. A 1D pointwise convolution is then applied to expand the channel dimension and produce preliminary weights. A softmax activation (applied along the channel dimension) yields the final fusion coefficients:
\begin{equation}
    \omega_j = \operatorname{softmax}\left(W_1 \cdot \mathrm{Ada_{mean}}(\mathbf{X}_j)\right) \in \mathbb{R}^{\frac{3C}{2} \times 1}
\end{equation}
Here, $W_1$ is the convolution weight, and $Ada_{mean}(\cdot)$ denotes adaptive average pooling. The resulting $\omega_j$ is partitioned along the channel dimension into
\begin{equation}
\omega_{j,i} \in \mathbb{R}^{\frac{C}{2}\times 1}, i \in \{0,1,2\},
\end{equation}
which are used to adaptively modulate the multi-scale feature maps.

The multi-scale features $Y_{j,i}$ obtained from parallel convolutions are then adaptively aggregated using the input-dependent weights $\omega_{j,i}$:
\begin{equation}
    F_j = \sum_{i=0}^{2} Y_{j,i} \odot \omega_{j,i} \in \mathbb{R}^{\frac{C}{2} \times T}
\end{equation}
where $\odot$ denotes channel-wise broadcast multiplication.

The outputs of the two MFE branches are concatenated and fused via a 1D convolution:
\begin{equation}
    \mathbf{X_{n}'} = W_2 \cdot \mathrm{Concat}(F_0,F_1) \in \mathbb{R}^{C \times T}
\end{equation}
where $Concat(\cdot)$ concatenates along the channel dimension, and $W_2$ is a 1D convolution for integrating the fused features.

Compared to traditional static fusion, this input-adaptive scheme dynamically adjusts aggregation weights according to the input features, achieving input-aware modeling with enhanced representational capacity for complex inertial odometry tasks.

\textbf{Adaptive Gating Unit (AGU).} To prevent the Multilayer Perceptron (MLP) in standard encoder architectures from disrupting the relative positional information in IMU measurements, we propose replacing the MLP with an AGU that adaptively gates local nearest-neighbor motion features. The AGU comprises two branches: a gating branch and a value branch.

\begin{figure*}[!t]
\centering
\captionsetup{aboveskip=2pt,font=small}
\includegraphics[width=0.95\textwidth]{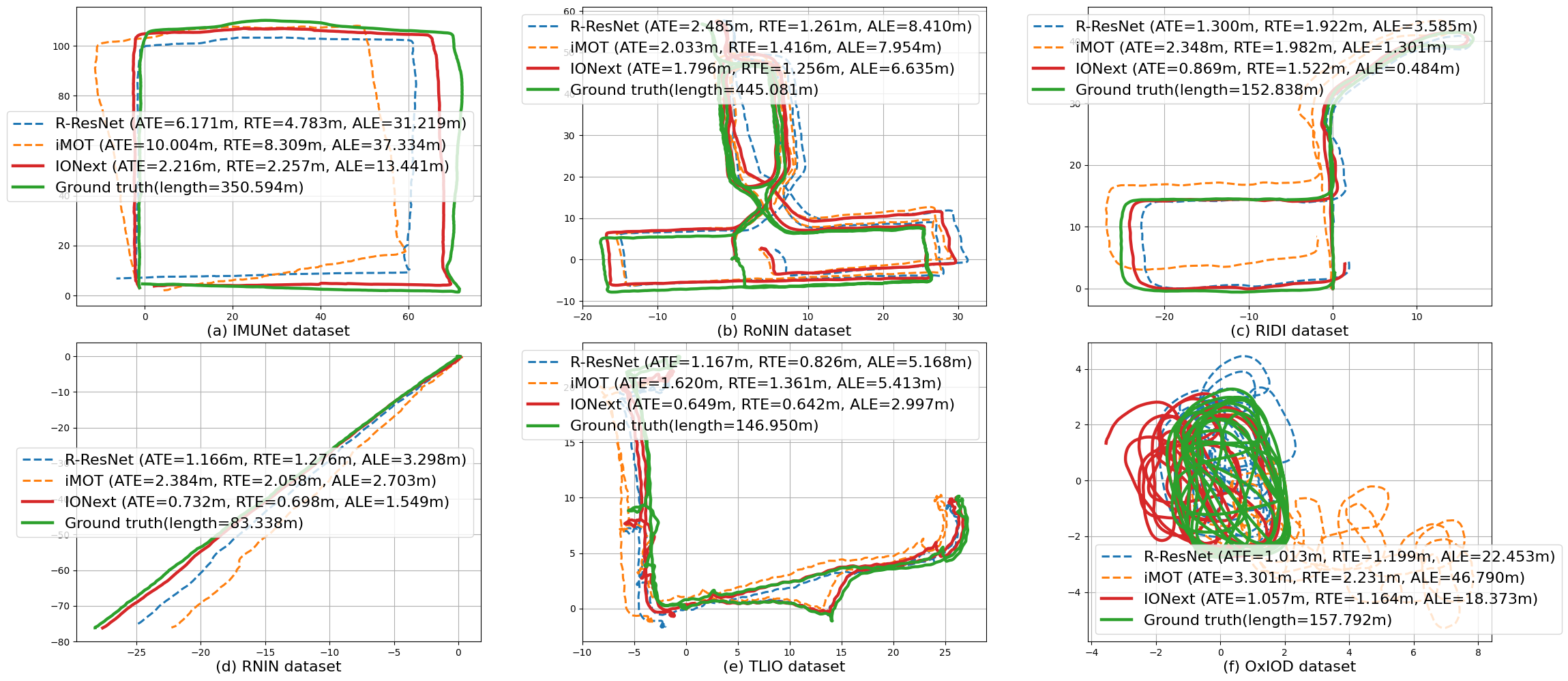}
\caption{Visualization of sample trajectories across six datasets, comparing IONext with two baseline models (R-ResNet and iMOT).}
\label{combined_plot}
\vspace{-20pt}
\end{figure*}

\textit{1) Gating branch.} To assess the per-channel contribution of IMU measurements, the gating branch applies global adaptive max pooling to emphasize transient motion changes and global adaptive average pooling to capture overall measurement context:
\begin{equation}
X_{\mathrm{mean}} = \operatorname{Ada}_{\mathrm{mean}}\bigl(X_n'\bigr) \in \mathbb{R}^{C\times 1}
\end{equation}
\begin{equation}
X_{\mathrm{max}} = \operatorname{Ada}_{\mathrm{max}}\bigl(X_n'\bigr) \in \mathbb{R}^{C\times 1}
\end{equation}
where \(\operatorname{Ada}_{\mathrm{mean}}\) and \(\operatorname{Ada}_{\mathrm{max}}\) denote global adaptive average pooling and global adaptive max pooling, respectively. The descriptors \(X_{\mathrm{mean}}\) and \(X_{\mathrm{max}}\) are complementary and enable multi-scale modeling within the AGU. The two descriptors are concatenated and fused by a 1-D convolution; a sigmoid activation \(\upsigma\) then produces the per-channel gating weights:
\begin{equation}
\upxi = \upsigma\!\Bigl(W_{3}\,\operatorname{Concat}\bigl(X_{\mathrm{mean}},\,X_{\mathrm{max}}\bigr)\Bigr) \in \mathbb{R}^{C\times 1}
\end{equation}
where $W_3$ represents the learnable weights of the 1D convolution. The resulting gating weights \(\upxi\) dynamically encode contextual information and fine-grained motion variations per channel.

\textit{2) Value branch.} We adopt lightweight depthwise convolutions to efficiently extract nearest-neighbor motion-variation features while preserving each token's temporal structure\cite{TransNeXt}. This design leverages local motion consistency and preserves the structural integrity of the IMU measurements.

The final output is obtained by elementwise multiplication of the gating weights and the value-branch features:
\begin{equation}
\mathbf{X_{n+1}} = \upxi \odot \operatorname{DWConv}\bigl(X_n'\bigr) \in \mathbb{R}^{C\times T}
\end{equation}
Through this mechanism, the AGU adaptively assigns dynamic importance weights to each channel and regulates the features of nearest-neighbor motion variations, thereby enhancing IONext's capability to model motion states.

\section{Experiments and Analysis}
\subsection{Experimental Settings}
\textbf{Datasets.} We conduct experiments on six publicly available benchmark datasets: IMUNet\cite{IMUNet}, RoNIN\cite{RoNIN}, RIDI\cite{RIDI}, OxIOD\cite{Oxiod}, RNIN\cite{RNIN-VIO}, and TLIO\cite{TLIO}. All datasets are randomly re-split into training, validation, and test subsets with a ratio of \(8:1:1\).

\textbf{Implementation Details.}
During training, we use the Adam optimizer with a batch size of 512 and a maximum of 100 epochs. The initial learning rate is set to \(10^{-4}\), and training is terminated early if the learning rate falls below \(10^{-6}\) to mitigate overfitting. All training and evaluation are performed on an NVIDIA RTX 3090 GPU with 24~GB of memory. 

\textbf{Baselines}
Recent studies have shown that data-driven inertial odometry methods significantly outperform traditional approaches based on Newtonian mechanics~\cite{surveyILS,Ionet,AIO,LIDR,NILoc}. Therefore, we select representative learning-based methods as baselines, including the R-LSTM\cite{LSTM} and R-ResNet\cite{ResNet} introduced in RoNIN\cite{RoNIN}, IMUNet\cite{IMUNet}, and the neural-network architectures proposed in TLIO\cite{TLIO} and RNIN\cite{RNIN-VIO}. In addition, we evaluate recent Transformer-based methods such as SBIPTVT~\cite{SBIPTVT}, CTIN\cite{CTIN}, and iMOT\cite{iMOT}.

\begin{figure*}[!t]
\centering
\captionsetup{aboveskip=2pt,font=small}
\includegraphics[width=1.0\textwidth]{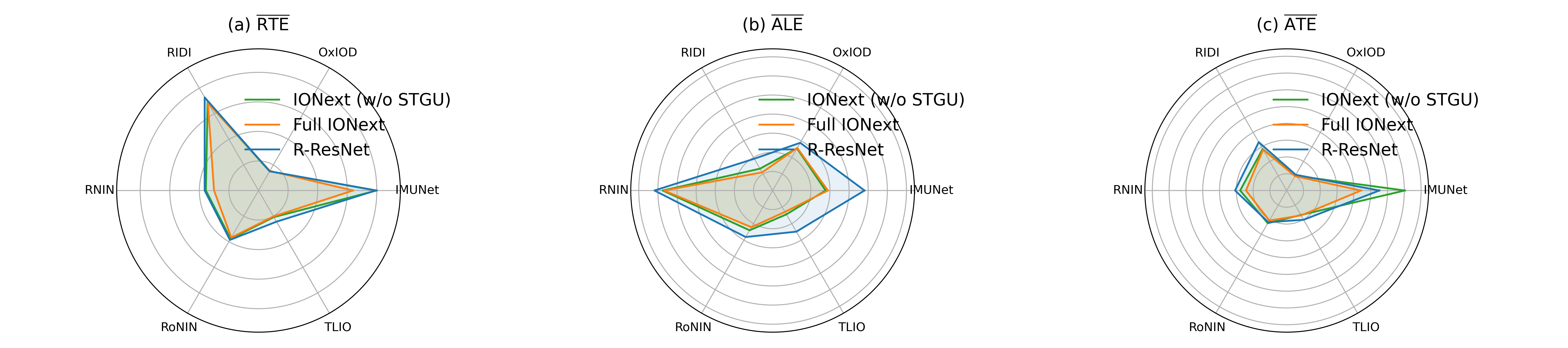}
\caption{(a), (b) and (c) are radar plots of $\overline{RTE}$, $\overline{ALE}$, and $\overline{ATE}$, respectively, for R-ResNet, IONext (w/o AGU), and the Full IONext on six benchmark datasets. Smaller polygon areas indicate lower errors.}
\label{radar_chart}
\vspace{-20pt}
\end{figure*}

\subsection{Trajectory Error Metrics and Normalization Strategy}
In the IO field, \(\mathrm{ATE}\)\cite{IDOL} and \(\mathrm{RTE}\)\cite{IDOL} are commonly used evaluation metrics: they assess trajectory consistency from global and local perspectives, respectively. However, these metrics primarily capture pointwise positional errors and do not necessarily reflect the overall quality of trajectory reconstruction. To mitigate this limitation, we introduce an additional metric:

\begin{itemize}[noitemsep, nolistsep, leftmargin=*]
    \item \textbf{Absolute Length Error (ALE)} quantifies the discrepancy between the predicted total trajectory length \(\hat{L}\) and the ground-truth length \(L\):
    \begin{equation}
        \mathrm{ALE} \;=\; \bigl\lvert \hat{L} - L \bigr\rvert
    \end{equation}
\end{itemize}

Researchers typically average metrics across multiple test sequences to obtain dataset-level results. However, simple averaging ignores variations in trajectory lengths across sequences. To compensate for this, we apply a length-based normalization to all metrics. The unified computation is
\begin{equation}
    \bar{m}
    \;=\;
    \sum_{i=1}^{N} \frac{L_i}{\sum_{j=1}^{N} L_j}\cdot\frac{m_i}{L_i}
    \;=\;
    \frac{\sum_{i=1}^{N} m_i}{\sum_{j=1}^{N} L_j}
\end{equation}
where \(L_i\) denotes the ground-truth length of the \(i\)-th trajectory, \(m_i \in \{\mathrm{ATE}_i,\ \mathrm{RTE}_i,\ \mathrm{ALE}_i\}\) denotes the corresponding metric for trajectory \(i\), and \(N\) is the number of trajectories. The ratio \(m_i/L_i\) is the length-normalized error for trajectory \(i\), while \(L_i/\sum_{j=1}^N L_j\) denotes its relative weight. The normalized metrics are written as \(\overline{\mathrm{ATE}}\), \(\overline{\mathrm{RTE}}\), and \(\overline{\mathrm{ALE}}\). 

This normalization ensures that each trajectory contributes proportionally to the overall metric according to its length, preventing short trajectories from disproportionately skewing the dataset-level results.

\subsection{Comparison with State-of-the-Art Methods}
\textbf{Quantitative Comparison.}
The quantitative evaluation across several benchmark datasets is summarized in Table~\ref{results}. Among data-driven inertial odometry methods, the proposed IONext outperforms alternatives, achieving the lowest errors in the majority of test scenarios. For example, on the RNIN dataset, IONext reduces \(\overline{\mathit{ATE}}\), \(\overline{\mathit{RTE}}\), and \(\overline{\mathit{ALE}}\) by \(1.01\,\mathrm{m}\), \(0.66\,\mathrm{m}\), and \(2.29\,\mathrm{m}\), respectively, relative to the current best-performing method iMOT. The improvement is even more pronounced relative to earlier CNN--based architectures (e.g., R-ResNet). The consistent advantage of IONext across six datasets indicates strong generalization and effective adaptation to diverse indoor and outdoor motion scenarios.

\textbf{Model Performance Analysis.}
Due to space limitations, we present results only for the RNIN dataset. Fig.~\ref{CDF_and_PDE}(a) and (b) plot the CDFs of \(\mathit{ATE}\) and \(\mathit{RTE}\), respectively. Notably, the red curve for IONext lies consistently toward the upper-left region of the plots, indicating lower trajectory errors and higher cumulative probability. For instance, IONext attains \(P(\mathit{ATE}<0.12\,\mathrm{m}) = 0.80\), i.e., \(80\%\) of its estimated trajectory points have \(\mathit{ATE}<0.12\,\mathrm{m}\). At the same probability level, the \(\mathit{ATE}\) values for R-ResNet (CNN--based), R-LSTM (LSTM--based), and iMOT (Transformer--based) are approximately \(0.13\,\mathrm{m}\), \(0.21\,\mathrm{m}\), and \(0.17\,\mathrm{m}\), respectively.

\textbf{Trajectory Reconstruction Visualization.}
Fig.~\ref{combined_plot} compares predicted trajectories of representative methods with the ground truth. R-ResNet (pure CNN) shows increasing deviation from the ground-truth trajectory as trajectory length grows. Although the Transformer--based iMOT mitigates some long-range errors, it still suffers from significant drift after multiple turns. In contrast, IONext produces trajectories that closely follow the ground truth, owing to its adaptive multi-scale modeling of motion. Consequently, IONext yields substantially improved trajectory reconstruction relative to the other methods.

\subsection{Ablation Study}
Our proposed IONext includes two key components, ADM and AGU. To quantify their contributions, we integrate them into two backbones (IONext and R-ResNet) to create several variants. Fig.~\ref{CDF_and_PDE}(c)--(d) show the CDFs of \( \mathit{ATE} \) and \( \mathit{RTE} \) on the RNIN dataset (due to space constraints), illustrating the effect of removing AGU from IONext and of progressively adding ADM and AGU to R-ResNet. The configuration with both ADM and AGU (red and orange curves) outperforms the variant with ADM only (green and blue curves), indicating that the two modules are complementary.
Fig.~\ref{radar_chart} further summarizes the ablation results on six datasets. IONext without AGU (green hexagon) still yields substantial gains over the R-ResNet baseline (blue hexagon), highlighting the effectiveness of ADM. Adding AGU (orange hexagon) brings additional improvements and yields the best localization accuracy among all configurations.

In summary, ADM and AGU each provide clear benefits across architectures, and their combination significantly improves localization accuracy.

\section{Conclusion}
We present IONext, which integrates Transformer-style architectural principles and multi-scale receptive fields into a convolutional framework. IONext adaptively aggregates contextual motion cues and fine-grained motion variation features from the input. Future work includes combining IONext with filtering-based methods to further improve localization accuracy, and evaluating its effectiveness on platforms such as ground vehicles and UAVs.

\bibliographystyle{IEEEtran}
\bibliography{refs}

\begin{thebibliography}{10}
\providecommand{\url}[1]{#1}
\csname url@rmstyle\endcsname
\providecommand{\newblock}{\relax}
\providecommand{\bibinfo}[2]{#2}
\providecommand\BIBentrySTDinterwordspacing{\spaceskip=0pt\relax}
\providecommand\BIBentryALTinterwordstretchfactor{4}
\providecommand\BIBentryALTinterwordspacing{\spaceskip=\fontdimen2\font plus
\BIBentryALTinterwordstretchfactor\fontdimen3\font minus \fontdimen4\font\relax}
\providecommand\BIBforeignlanguage[2]{{%
\expandafter\ifx\csname l@#1\endcsname\relax
\typeout{** WARNING: IEEEtran.bst: No hyphenation pattern has been}%
\typeout{** loaded for the language `#1'. Using the pattern for}%
\typeout{** the default language instead.}%
\else
\language=\csname l@#1\endcsname
\fi
#2}}

\bibitem{AirIO}
Y.~Qiu, C.~Xu, Y.~Chen, S.~Zhao, J.~Geng, and S.~Scherer, ``Airio: Learning inertial odometry with enhanced imu feature observability,'' \emph{IEEE Robotics and Automation Letters}, pp. 1--8, 2025.

\bibitem{Tartan-IMU}
\BIBentryALTinterwordspacing
S.~Zhao, S.~Zhou, R.~Blanchard, Y.~Qiu, W.~Wang, and S.~Scherer, ``Tartan imu: A light foundation model for inertial positioning in robotics,'' \emph{2025 IEEE/CVF Conference on Computer Vision and Pattern Recognition (CVPR)}, pp. 22\,520--22\,529, 2025. [Online]. Available: \url{https://api.semanticscholar.org/CorpusID:280089012}
\BIBentrySTDinterwordspacing

\bibitem{RIO}
X.~Cao, C.~Zhou, D.~Zeng, and Y.~Wang, ``Rio: Rotation-equivariance supervised learning of robust inertial odometry,'' in \emph{CVPR}, 2022, pp. 6604--6613.

\bibitem{SurveyofIndoorInertial}
R.~Harle, ``A survey of indoor inertial positioning systems for pedestrians,'' \emph{IEEE Communications Surveys \& Tutorials}, vol.~15, no.~3, pp. 1281--1293, 2013.

\bibitem{SINS}
P.~G. Savage, ``Strapdown inertial navigation integration algorithm design part 2: Velocity and position algorithms,'' \emph{Journal of Guidance, Control, and Dynamics}, vol.~21, no.~2, pp. 208--221, 1998.

\bibitem{PDRusingfrequencydomain}
M.~Kourogi and T.~Kurata, ``A method of pedestrian dead reckoning for smartphones using frequency domain analysis on patterns of acceleration and angular velocity,'' in \emph{2014 IEEE/ION Position, Location and Navigation Symposium-PLANS 2014}.\hskip 1em plus 0.5em minus 0.4em\relax IEEE, 2014, pp. 164--168.

\bibitem{AdaptiveThreshold-BasedZUPT}
H.~Li, H.~Liu, Z.~Li, C.~Li, Z.~Meng, N.~Gao, and Z.~Zhang, ``Adaptive threshold-based zupt for single imu-enabled wearable pedestrian localization,'' \emph{IEEE Internet of Things Journal}, vol.~10, no.~13, pp. 11\,749--11\,760, 2023.

\bibitem{IMU-and-magnetometer-modeling-for-smartphone-based-PDR}
R.~Hostettler and S.~Särkkä, ``Imu and magnetometer modeling for smartphone-based pdr,'' in \emph{2016 International Conference on Indoor Positioning and Indoor Navigation (IPIN)}, 2016, pp. 1--8.

\bibitem{RoNIN}
S.~Herath, H.~Yan, and Y.~Furukawa, ``Ronin: Robust neural inertial navigation in the wild: Benchmark, evaluations, \& new methods,'' in \emph{2020 IEEE International Conference on Robotics and Automation (ICRA)}, 2020, pp. 3146--3152.

\bibitem{IMUNet}
B.~Zeinali, H.~Zanddizari, and M.~J. Chang, ``Imunet: Efficient regression architecture for inertial imu navigation and positioning,'' \emph{IEEE Transactions on Instrumentation and Measurement}, vol.~73, no. 2516213, 2024.

\bibitem{InceptionTransformer}
C.~Si, W.~Yu, P.~Zhou, Y.~Zhou, X.~Wang, and S.~Yan, ``Inception transformer,'' in \emph{Proceedings of the 36th International Conference on Neural Information Processing Systems}, ser. NIPS '22.\hskip 1em plus 0.5em minus 0.4em\relax Red Hook, NY, USA: Curran Associates Inc., 2022.

\bibitem{Attentionisallyouneed}
A.~Vaswani, N.~Shazeer, N.~Parmar, J.~Uszkoreit, L.~Jones, A.~N. Gomez, L.~Kaiser, and I.~Polosukhin, ``Attention is all you need,'' in \emph{Proceedings of the 31st International Conference on Neural Information Processing Systems}, ser. NIPS'17.\hskip 1em plus 0.5em minus 0.4em\relax Red Hook, NY, USA: Curran Associates Inc., 2017, p. 6000–6010.

\bibitem{TKSA}
X.~Chen, H.~Li, M.~Li, and J.~Pan, ``Learning a sparse transformer network for effective image deraining,'' in \emph{Proceedings of the IEEE/CVF Conference on Computer Vision and Pattern Recognition (CVPR)}, June 2023, pp. 5896--5905.

\bibitem{DeepILS}
O.~Tariq, B.~Dastagir, M.~Bilal, and D.~Han, ``Deepils: Towards accurate domain invariant aiot-enabled inertial localization system,'' \emph{IEEE Internet of Things Journal}, pp. 1--1, 2025.

\bibitem{CTIN}
\BIBentryALTinterwordspacing
B.~Rao, E.~Kazemi, Y.~Ding, D.~M. Shila, F.~M. Tucker, and L.~Wang, ``Ctin: Robust contextual transformer network for inertial navigation,'' \emph{Proceedings of the AAAI Conference on Artificial Intelligence}, vol.~36, no.~5, pp. 5413--5421, Jun. 2022. [Online]. Available: \url{https://ojs.aaai.org/index.php/AAAI/article/view/20479}
\BIBentrySTDinterwordspacing

\bibitem{iMOT}
S.~M. {Nguyen}, L.~D. {Tran}, D.~{Viet Le}, and P.~J.~M. {Havinga}, ``{iMoT: Inertial Motion Transformer for Inertial Navigation},'' \emph{arXiv e-prints}, p. arXiv:2412.12190, Dec. 2024.

\bibitem{ConvNext}
Z.~Liu, H.~Mao, C.-Y. Wu, C.~Feichtenhofer, T.~Darrell, and S.~Xie, ``A convnet for the 2020s,'' \emph{Proceedings of the IEEE/CVF Conference on Computer Vision and Pattern Recognition (CVPR)}, 2022.

\bibitem{RepLKNet}
X.~Ding, X.~Zhang, J.~Han, and G.~Ding, ``Scaling up your kernels to 31×31: Revisiting large kernel design in cnns,'' in \emph{2022 IEEE/CVF Conference on Computer Vision and Pattern Recognition (CVPR)}, 2022, pp. 11\,953--11\,965.

\bibitem{AlexNet}
\BIBentryALTinterwordspacing
A.~Krizhevsky, I.~Sutskever, and G.~E. Hinton, ``Imagenet classification with deep convolutional neural networks,'' \emph{Commun. ACM}, vol.~60, no.~6, p. 84–90, May 2017. [Online]. Available: \url{https://doi.org/10.1145/3065386}
\BIBentrySTDinterwordspacing

\bibitem{Inceptionv1}
C.~Szegedy, W.~Liu, Y.~Jia, P.~Sermanet, S.~Reed, D.~Anguelov, D.~Erhan, V.~Vanhoucke, and A.~Rabinovich, ``Going deeper with convolutions,'' in \emph{2015 IEEE Conference on Computer Vision and Pattern Recognition (CVPR)}, 2015, pp. 1--9.

\bibitem{Inceptionnext}
W.~Yu, P.~Zhou, S.~Yan, and X.~Wang, ``Inceptionnext: When inception meets convnext,'' in \emph{Proceedings of the IEEE/CVF Conference on Computer Vision and Pattern Recognition}, 2024, pp. 5672--5683.

\bibitem{TransXNet}
M.~Lou, S.~Zhang, H.-Y. Zhou, S.~Yang, C.~Wu, and Y.~Yu, ``Transxnet: Learning both global and local dynamics with a dual dynamic token mixer for visual recognition,'' \emph{IEEE Transactions on Neural Networks and Learning Systems}, 2025.

\bibitem{RIDI}
H.~Yan, Q.~Shan, and Y.~Furukawa, ``Ridi: Robust imu double integration,'' in \emph{ECCV}, V.~Ferrari, M.~Hebert, C.~Sminchisescu, and Y.~Weiss, Eds.\hskip 1em plus 0.5em minus 0.4em\relax Cham: Springer International Publishing, 2018, pp. 641--656.

\bibitem{PDRNet}
O.~Asraf, F.~Shama, and I.~Klein, ``Pdrnet: A deep-learning pedestrian dead reckoning framework,'' \emph{IEEE Sensors Journal}, vol.~22, no.~6, pp. 4932--4939, 2022.

\bibitem{Ionet}
C.~Chen, X.~Lu, A.~Markham, and N.~Trigoni, ``Ionet: Learning to cure the curse of drift in inertial odometry,'' in \emph{Proceedings of the AAAI Conference on Artificial Intelligence}, vol.~32, no.~1, 2018.

\bibitem{TLIO}
W.~Liu, D.~Caruso, E.~Ilg, J.~Dong, A.~I. Mourikis, K.~Daniilidis, V.~Kumar, and J.~Engel, ``Tlio: Tight learned inertial odometry,'' \emph{IEEE Robotics and Automation Letters}, vol.~5, no.~4, pp. 5653--5660, 2020.

\bibitem{LIDR}
D.~Yang, H.~Liu, X.~Jin, J.~Chen, C.~Wang, X.~Ding, and K.~Xu, ``Enhancing vio robustness under sudden lighting variation: A learning-based imu dead-reckoning for uav localization,'' \emph{IEEE Robotics and Automation Letters}, vol.~9, no.~5, pp. 4535--4542, 2024.

\bibitem{WDSNet}
\BIBentryALTinterwordspacing
Y.~Wang and Y.~Zhao, ``Wavelet dynamic selection network for inertial sensor signal enhancement,'' \emph{Proceedings of the AAAI Conference on Artificial Intelligence}, vol.~38, no.~14, pp. 15\,680--15\,688, Mar. 2024. [Online]. Available: \url{https://ojs.aaai.org/index.php/AAAI/article/view/29496}
\BIBentrySTDinterwordspacing

\bibitem{EqNIO}
\BIBentryALTinterwordspacing
R.~K. Jayanth, Y.~Xu, Z.~Wang, E.~Chatzipantazis, K.~Daniilidis, and D.~Gehrig, ``Eq{NIO}: Subequivariant neural inertial odometry,'' in \emph{The Thirteenth International Conference on Learning Representations}, 2025. [Online]. Available: \url{https://openreview.net/forum?id=C8jXEugWkq}
\BIBentrySTDinterwordspacing

\bibitem{NILoc}
S.~Herath, D.~Caruso, C.~Liu, Y.~Chen, and Y.~Furukawa, ``Neural inertial localization,'' in \emph{2022 IEEE/CVF Conference on Computer Vision and Pattern Recognition (CVPR)}, 2022, pp. 6594--6603.

\bibitem{SBIPTVT}
X.~Li, K.~Li, J.~Liu, and R.~Gao, ``Smartphone-based indoor pedestrian tracking via transformer,'' in \emph{2024 27th International Conference on Computer Supported Cooperative Work in Design (CSCWD)}, 2024, pp. 1280--1285.

\bibitem{Swin-ransformer-V2}
Z.~Liu, H.~Hu, Y.~Lin, Z.~Yao, Z.~Xie, Y.~Wei, J.~Ning, Y.~Cao, Z.~Zhang, L.~Dong, F.~Wei, and B.~Guo, ``Swin transformer v2: Scaling up capacity and resolution,'' in \emph{International Conference on Computer Vision and Pattern Recognition (CVPR)}, 2022.

\bibitem{Inceptionv3}
C.~Szegedy, V.~Vanhoucke, S.~Ioffe, J.~Shlens, and Z.~Wojna, ``Rethinking the inception architecture for computer vision,'' in \emph{2016 IEEE Conference on Computer Vision and Pattern Recognition (CVPR)}, 2016, pp. 2818--2826.

\bibitem{SLaK}
S.~Liu, T.~Chen, X.~Chen, X.~Chen, Q.~Xiao, B.~Wu, M.~Pechenizkiy, D.~Mocanu, and Z.~Wang, ``More convnets in the 2020s: Scaling up kernels beyond 51x51 using sparsity,'' \emph{arXiv preprint arXiv:2207.03620}, 2022.

\bibitem{Local-Attention-and-Dynamic-Depth-wise-Convolution}
Q.~Han, Z.~Fan, Q.~Dai, L.~Sun, M.-M. Cheng, J.~Liu, and J.~Wang, ``On the connection between local attention and dynamic depth-wise convolution,'' in \emph{International Conference on Learning Representations}, 2022.

\bibitem{TransNeXt}
D.~Shi, ``Transnext: Robust foveal visual perception for vision transformers,'' in \emph{Proceedings of the IEEE/CVF Conference on Computer Vision and Pattern Recognition (CVPR)}, June 2024, pp. 17\,773--17\,783.

\bibitem{Oxiod}
C.~Chen, P.~Zhao, C.~X. Lu, W.~Wang, A.~Markham, and N.~Trigoni, ``Oxiod: The dataset for deep inertial odometry,'' \emph{arXiv preprint arXiv:1809.07491}, 2018.

\bibitem{RNIN-VIO}
D.~Chen, N.~Wang, R.~Xu, W.~Xie, H.~Bao, and G.~Zhang, ``Rnin-vio: Robust neural inertial navigation aided visual-inertial odometry in challenging scenes,'' in \emph{2021 IEEE International Symposium on Mixed and Augmented Reality (ISMAR)}, 2021, pp. 275--283.

\bibitem{surveyILS}
A.~K. Panja, C.~Chowdhury, and S.~Neogy, ``Survey on inertial sensor-based ils for smartphone users,'' \emph{CCF Transactions on Pervasive Computing and Interaction}, vol.~4, no.~3, pp. 319--337, 2022.

\bibitem{AIO}
R.~Fu, Q.~Niu, M.~Kou, Q.~He, and N.~Liu, ``Aio: Enhancing generalization of inertial odometry to unseen scenarios with active learning,'' in \emph{2024 International Symposium on Digital Home (ISDH)}, 2024, pp. 155--160.

\bibitem{LSTM}
S.~Hochreiter and J.~Schmidhuber, ``Long short-term memory,'' \emph{Neural Computation}, vol.~9, no.~8, pp. 1735--1780, 1997.

\bibitem{ResNet}
K.~He, X.~Zhang, S.~Ren, and J.~Sun, ``Deep residual learning for image recognition,'' in \emph{Proceedings of the 2016 IEEE Conference on Computer Vision and Pattern Recognition (CVPR)}, 2016, pp. 770--778.

\bibitem{IDOL}
C.~L. Gentil, F.~Tschopp, I.~Alzugaray, T.~Vidal-Calleja, R.~Siegwart, and J.~Nieto, ``Idol: A framework for imu-dvs odometry using lines,'' 2020, arXiv:2008.05749.

\end{thebibliography}

\end{document}